# Unsupervised Disentangled Representation Learning with Analogical Relations

**Zejian Li, Yongchuan Tang** *, **Yongxing He,**
College of Computer Science, Zhejiang University, Hangzhou 310027, China
{zejianlee, yctang, heyongxing}@zju.edu.cn,

## Abstract

Learning the disentangled representation of interpretable generative factors of data is one of the foundations to allow artificial intelligence to think like people. In this paper, we propose the analogical training strategy for the unsupervised disentangled representation learning in generative models. The analogy is one of the typical cognitive processes, and our proposed strategy is based on the observation that sample pairs in which one is different from the other in one specific generative factor show the same analogical relation. Thus, the generator is trained to generate sample pairs from which a designed classifier can identify the underlying analogical relation. In addition, we propose a disentanglement metric called the subspace score, which is inspired by subspace learning methods and does not require supervised information. Experiments show that our proposed training strategy allows the generative models to find the disentangled factors, and that our methods can give competitive performances as compared with the state-of-the-art methods.

## 1 Introduction

This paper is concerned with the problem of unsupervised disentangled representation learning in the generative model. The disentangled representation is a kind of distributed feature representation in which disjoint dimensions of a latent code reflect different high-level generative factors of data. Specifically, the disentangled representation can separate the explanatory factors which interact nonlinearly in the real-world data, such as the object shape, the material property, and the light source, etc. Thus, the disentangled representation is helpful for a large variety of AI tasks [Bengio *et al.*, 2013].

The generative model is helpful in learning the disentangled representation. It is a methodology to learn a probability distribution, and it generates a new sample according to the code in the hidden space. By learning the appropriate parameter, it can gradually learn to generate new data of the same distribution as the target one [Goodfellow *et al.*, 2014].

When the disentangled representation is learned in the generative model, disjoint dimensions of a hidden code could model the data generative factors separately. These underlying factors could explain the major variation in the data. When only one factor varies but all others are fixed, the generated sequence of samples can show an interpretable change to human beings. For example, when we generate a figure of a hand-written digital number, a component of the code may be associated with the stroke width. When the value is changed, the stroke width of the generated number becomes smaller. A large body of work has been devoted to this problem. When the generative factors are predefined, the disentangled representation can be learned by the reconstruction of the data when the codes are swapped [Peng *et al.*, 2017; Denton and Birodkar, 2017]. Moreover, when the data are labeled with attributes, the representation can be learned via the mapping between the data and the attributes [Wang *et al.*, 2017] or the consistency between the variation of the data and the transformation of the latent code [Kulkarni *et al.*, 2015; Worrall *et al.*, 2017]. On the other hand, unsupervised learning of a disentangled representation has been a major challenge. DIP-VAE [Higgins *et al.*, 2017] and $\beta$-VAE [Kumar *et al.*, 2017] learn the disentanglement of the latent code by encouraging the latent distribution to be close to the standard normal distribution, in which each random variables are independent. InfoGAN [Chen *et al.*, 2016] also uses the statistical independence and the method is motivated by the principle of the maximization of the mutual information. In summary, most of the existing works disentangle the factors by taking advantage of the supervised signals or by using the statistical independence of the prior distribution.

Different from the existing methods, our solution is motivated by the analogy. A key observation is that each interpretable disentangled factor is associated with an analogical relation of sample pairs. In the example of generating digital numbers, disjoint components of a hidden code can be associated with the factors such as rotation, stroke thickness, width, etc. Figure 1 gives an illustrative example of analogical pairs, which manifests the factor of stroke thickness. Reversely, this factor of variation can be learned from the analogical sample pairs. The discussed relation is known as the proportional analogical relation, which has a general form $(a : b :: c : d)$ where the pairs $(a, b)$ and $(c, d)$ have relational similarity, such as mammals : lungs :: fish : gills [Gust *et al.*, 2008]. The analogical relation can be extended to a batch of $n$ sample pairs, which have the form $(a_1 : b_1 :: a_2 : b_2 :: \ldots :: a_n : b_n)$. Namely, the pairs of $a_i$ and $b_i$ for all $i = 1, \ldots, n$ share the same analogical relation. The analogy has been a central part of human intelligence and cognition, so from this perspective learning the interpretable factors via the analogical relation is close to the human-like cognition process.

---
*Corresponding author

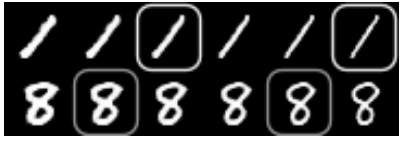

Figure 1: **An illustrative example of the analogical relation.** These hand-written numbers are generated by our proposed Analogical GAN method trained on the MNIST dataset. The stroke becomes thinner from left to right. A pair of "1"s and a pair of "8"s are selected, which show the analogical relation of stroke thickness. The relation of the first bolder "1" to the second thinner "1" is similar to that of the first bolder "8" to the second thinner "8". Furthermore, the "1"s here are rotated while "8"s are not. Given this difference in rotation, the variation of stroke width shows the same pattern.

Based on the observation above, we propose our analogical training strategy on top of the generative model. At the beginning of the training process, the generator gives an analogical sample pair according to the code pair in which a predefined component of the latent code is different while all the other configuration is fixed. Next, an extra classifier tries to identify the predefined analogical relation behind the generated analogical pair. Then, the classifier and the generator are trained together to allow the classifier to make the correct decision. This is a cooperative game. The generator should learn to generate sample pairs characterized by analogical relations which the classifier can capture. Figure 2 gives an illustration of the analogical training process. This is our main contribution.

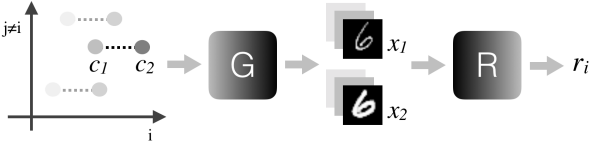

Figure 2: **The analogical training process.** We assume each code component represents a generative factor, and each generative factor has a unique analogical relation. The analogical relation $r_i$ of the factor is implied by a pair of codes which are different only in the $i$th component. In the left of the figure, the horizontal axis is the $i$th dimension, and the vertical dimension indicates all the other dimensions. The code pair $c_1$ and $c_2$ are different only in the $i$th one. To start the training process, the code pair is passed through the generator $G$, which generates the analogical sample pair $x_1$ and $x_2$. Then, the classifier $R$ tries to identify the underlying relation $r_i$. Finally, the generator and the classifier both update their parameters such that the analogical relation can be classified more accurately, and thus the $i$th component learns to reflect a meaningful factor gradually. The training process can be applied on a batch of code pairs, and the analogical sample pairs also share the relation $r_i$.

The other contribution is our proposal of a disentanglement metric of the learned representation, called the subspace score.

This metric is based on two assumptions inspired by the subspace learning methodology. The first assumption is that each kind of variation of the generated samples forms an affine subspace, and thus these subspaces are expected to be uncovered by the subspace clustering algorithm. Therefore, the disentanglement of the factors is approximated by the clustering performance. The second is that the union of these subspaces should be close to the majority of observed samples. This closeness is measured by the distance between the observed samples and the affine space spanned by the generated samples. Our proposed subspace score is the combination of these two measures. The subspace score does not require supervised information, and it is able to be applied to the real-world unlabeled dataset. To the best of our knowledge, this is the first time an unsupervised disentanglement metric is proposed.

The rest of the paper is organized as follows. We introduce our proposed analogical training strategy in Section 2 and the disentanglement metric in Section 3. Next, we review related works in Section 4. In Section 5 we demonstrate the experiment results and compare our method with other methods along the subspace score. We conclude the paper in Section 6. Our source code will be available on https://github.com/ZejianLi/analogical-training.

## 2 Method
### 2.1 Preliminaries

In this part, we briefly review two generative models, Variational Auto-Encoder (VAE) [Kingma and Welling, 2013] and Generative Adversarial Network (GAN) [Goodfellow et al., 2014]. A generative model learns the target probability distribution by generating new samples whose distribution is close the target one. Formally, given a latent representation $z \in \mathcal{Z}$ sampled from a predefined distribution $\mathbb{P}_z$, we can generate a new sample $x \in \mathcal{X}$ by sampling from $\mathbb{P}_\theta(x \mid z)$. Here $\mathbb{P}_\theta(x \mid z)$ is assumed to be deterministic and described by a function $G \colon \mathcal{Z} \mapsto \mathcal{X}$ parameterized by $\theta$, and thus $x = G(z)$. By learning appropriate $\theta$, the generated distribution $\mathbb{P}_\theta$ can get closer to the ground truth distribution $\mathbb{P}_r$.

VAE has emerged as a popular deep generative model. A key step in VAE is to reinterpret the log-likelihood of the observed sample $x$ in the generated distribution $\mathbb{P}_\theta$ as

$$\log p_\theta(x) = \mathrm{KL}(q_\phi(z \mid x) \| p_\theta(z \mid x)) + \mathcal{L}(\theta, \phi; x). \quad (1)$$

Here $q_\phi(z \mid x)$ is the variational posterior distribution with the parameter $\phi$ and $\mathrm{KL}(\cdot \| \cdot)$ is the Kullback-Leibler divergence between two distributions. $\mathcal{L}(\theta, \phi; x)$ is the evidence lower bound defined as

$$\mathcal{L}(\theta, \phi; x) = \mathbb{E}_{q_\phi(z|x)} \log p_\theta(x \mid z) - \mathrm{KL}(q_\phi(z \mid x) \| p(z)). \quad (2)$$

The first term is the expected log-likelihood to recover $x$, and the second term is the KL-divergence between the variational posterior distribution and the prior distribution. Since we always have $\log p_\theta(x) \geq \mathcal{L}(\theta, \phi; x)$, we can increase the log-likelihood of the data by maximizing the evidence lower bound. Hence, the optimization problem of the model is

$$\begin{aligned} \max_{G,Q} \quad & L(G, Q) \\ \text{s.t.} \quad & L(G, Q) = \mathbb{E}_{x \sim \mathbb{P}_r} \mathcal{L}(\theta, \phi; x). \end{aligned} \quad (3)$$

Here we use the function $Q$ parameterized by $\phi$ to describe $q_\phi(z \mid x)$.

GAN is another framework to train generative models. It learns the real distribution by training the generator $G$ to confuse an adversarial discriminator $D$. The discriminator $D$ tries to distinguish the data generated from $\mathbb{P}_\theta$ or sampled from $\mathbb{P}_r$, while the generator $G$ gradually learns to generate data that $D$ cannot correctly classify. Formally, given a sample $x$, $D(x)$ approximates the probability that $x$ is sampled from $\mathbb{P}_r$. The formulation is given as follows.

$$\min_G \max_D V(D,G)$$
$$\text{s.t.} V(D,G) = \mathbb{E}_{x \sim \mathbb{P}_r} \log D(x) + \mathbb{E}_{z \sim \mathbb{P}_z} \log(1 - D(G(z))) \quad (4)$$

To learn the disentangled representation in GAN, we divide the latent representation to the continuous code $c$ and the noise $z$. The latent code $c$ tries to capture the disentangled representation while the noise $z$ fits other details. Thus the generated sample is given by $x = G(c, z)$ where $c \sim \mathbb{P}_c$ and $z \sim \mathbb{P}_z$. We do not adopt this division in VAE, since we find it leads to a trivial solution. For consistency, we write $x = G(c, z)$ in the rest of the paper.

## 2.2 Analogical Training Strategy

In this part, we describe our proposed analogical training strategy, which is to learn the disentangled representation of interpretable factors. Our proposed method is based on the observation that a disentangled generative factor is related to a unique analogical relation of sample pairs. As shown in Figure 1, we have two different digits, and by reducing their stroke width we have their twins. The relation of the first digit and its twin is similar to the relation between the other digit and its twin. Even though the configurations of other factors are different, the change of the given factor shows the variation of the same pattern. Thus, the stroke width factor is shown by the analogical relation. Reversely, given a generator, if the analogical relation behind different generated pairs can always be recognized, we believe the learned factor is disentangled. More detailedly, each component of the hidden code represents a generative factor, and the change of the generative factor can be reflected by the variation in samples. Sample pairs which are different only in a single generative factor are defined as analogical pairs, and together they show the analogical relation unique to the generative factor. The model learns the disentangled factor by trying to generate analogical pairs.

Formally, we define a classifier $R$ to recognize the analogical relations. Given an analogical pair of samples $x_1$ and $x_2$, $R$ identifies the generative factor $r$ in which the two samples are different. This is to maximize the log-likelihood

$$\mathbb{E}_r \mathbb{E}_{x_1,x_2 \mid r} \log R(r \mid x_1, x_2) \quad (5)$$

in which $r$ is the generative factor as a random variable of the category distribution over $\{1, \ldots, k\}$ when we have $k$ factors. $\mathbb{E}_{x_1,x_2 \mid r}$ is short for $\mathbb{E}_{(x_1,x_2) \sim \mathbb{P}(x_1,x_2 \mid r)}$. Particularly, $R(r \mid x_1, x_2)$ is the probability that $R$ believe $x_1$ and $x_2$ are different in the factor $r$, which means $x_1$ and $x_2$ show the analogical relation unique to $r$.

Since samples in an analogical pair are different in one factor, we can generate the pair according to two latent codes which are different in one component. We denote the code pair as $c_1$ and $c_2$ and thus $x_1 = G(c_1, z)$ and $x_2 = G(c_2, z)$. Thus, (5) can be rewritten as

$$K(G,R) = \mathbb{E}_r \mathbb{E}_{c_1,c_2 \mid r} \mathbb{E}_{x_1 \sim G(c_1,z), x_2 \sim G(c_2,z)} \log R(r \mid x_1, x_2). \quad (6)$$

The process of the analogical training is visualized in Figure 2. The expectation can be approximated with the Monte Carlo method.

Theoretically, $K(G,R)$ is the lower bound of the mutual information $I(r; x_1, x_2)$ between the generative factor and the sample pair. Formally, $I(r; x_1, x_2) = H(r) - H(r \mid x_1, x_2)$, where the first term is the entropy of $r$, and the second term is the conditional entropy of $r$ given $x_1$ and $x_2$. We have $H(r) = \log k$ and

$$-H(r \mid x_1, x_2) = \mathbb{E}_{x_1,x_2} \mathbb{E}_{r \mid x_1,x_2} \log P(r \mid x_1, x_2). \quad (7)$$

The computation of the posterior distribution $P(r \mid x_1, x_2)$ is intractable. We can define an auxiliary distribution $R$ to infer a variational lower bound.

$$\begin{aligned}
&- H(r \mid x_1, x_2) \\
&= \mathbb{E}_{x_1,x_2} \mathbb{E}_{r \mid x_1,x_2} \left[ \log \frac{P(r \mid x_1, x_2)}{R(r \mid x_1, x_2)} + \log R(r \mid x_1, x_2) \right] \\
&= \mathbb{E}_{x_1,x_2} \text{KL}(P(r \mid x_1, x_2) \| R(r \mid x_1, x_2)) \\
&\quad + \mathbb{E}_r \mathbb{E}_{x_1,x_2 \mid r} \log R(r \mid x_1, x_2).
\end{aligned} \quad (8)$$

Since the KL-divergence is non-negative, we can have

$$\begin{aligned}
&- H(r \mid x_1, x_2) \\
&\geq \mathbb{E}_r \mathbb{E}_{x_1,x_2 \mid r} \log R(r \mid x_1, x_2) \\
&= \mathbb{E}_r \mathbb{E}_{c_1,c_2 \mid r} \mathbb{E}_{x_1 \sim G(z,c_1), x_2 \sim G(z,c_2)} \log R(r \mid x_1, x_2).
\end{aligned} \quad (9)$$

Therefore, $K(G,R)$ is the lower bound of $I(r; x_1, x_2)$. The bound is tight when $R(r \mid x_1, x_2)$ is close to $P(r \mid x_1, x_2)$. In this case, maximizing $K(G,R)$ is equivalent to maximizing the mutual information $I(r; x_1, x_2)$. Intuitively, this means the generator should make an effort to give sample pairs whose inner difference can show the variation of the generative factor.

Combining (3) and (6), we have our proposed Analogical VAE (AnaVAE) learned via the optimization problem

$$\max_{G,Q,R} L(G,Q) + \lambda K(G,R), \quad (10)$$

where $\lambda$ is the hyperparameter to control the effect of $K(G,R)$. Similarly, the optimization problem of our Analogical GAN (AnaGAN) is

$$\min_{G,R} \max_D V(D,G) - \lambda K(G,R). \quad (11)$$

Empirically, we set $\lambda$ to 1 by default. All the functions $G$, $Q$, $D$ and $R$ are modeled by neural networks.

## 3 Disentanglement Metric

In this part, we introduce our disentanglement metric. The metric is inspired by the subspace clustering algorithm [Elhamifar and Vidal, 2013]. We assume that sequences of samples of the same variation lie in a low-dimensional affine subspace. Then if the factors in the representation are disentangled, the subspaces of different variations are independent.[1] In this case, samples of the same variation can be grouped by the subspace clustering method. An exemplary illustration of subspaces of variations is given in Figure 3. Another assumption is that the observed samples should be close to the affine space spanned by the generated samples, because the generated samples should be indistinguishable to the real ones. Our metric is designed according to these two assumptions. Since these assumptions are not related to any supervised information, the subspace score can be applied to unlabeled datasets.

---

[1] A set of $k$ linear subspaces $\{S_i \subset \mathbb{R}^D\}_{i=1}^n$ are independent if $\dim(\oplus_{i=1}^k S_i) = \sum_{i=1}^k \dim(S_i)$, where $\oplus$ is the direct sum.

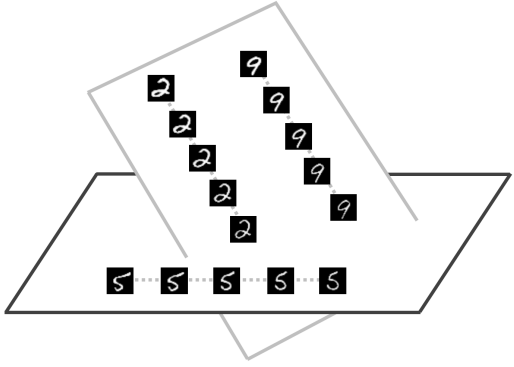

Figure 3: **An illustration of the subspaces of variations.** The samples of number "2" and "9" vary in the stroke width, and those of "5" changes in the writing style. We assume the sequences of the same variation lie on an affine subspace. Here, the sequences of "2" and "9" lie on the subspace of stroke variation, and the sequence "5" lies on the subspace of writing style variation. For visualization the two affine planes are not independent here.

The subspace clustering algorithm can separate data according to the relation of linear combinations and thus recover the low-dimensional affine subspaces. Given our assumption above, the disentanglement of factors can be approximated by how well the generated clusters of different variations can be correctly separated by the subspace clustering method. To be more specific, given the learned generator factor $r_i$, we generate a sequence of samples $\{x_j \mid j = 1, \ldots, m\}$ by varying the $i$th component of the code while all other configurations are fixed. Thus, this sequence shows the variation of $r_i$. A sample cluster consists of several different sequences of the same variation, and $k$ clusters of different variations are sampled and given to the subspace clustering algorithm. Formally, suppose we have $\mathbf{Y} = [\mathbf{Y_1}, \ldots, \mathbf{Y_k}]$. For $i \in \{1, \ldots, k\}$, $\mathbf{Y_i}$ is a group of sample sequences and the samples in each sequence are different only in the generative factor $r_i$. Then the coefficient matrix $\mathbf{U}$ is learned by

$$\hat{\mathbf{U}} = \arg\min \quad \|\mathbf{YU} - \mathbf{Y}\|_{\mathbf{F}}^{\mathbf{2}} + \lambda' \mathcal{R}(\mathbf{U}), \quad (12)$$
$$\text{s.t.} \quad \text{diag}(\mathbf{U}) = \mathbf{0}.$$

Here, $\|\cdot\|_F$ is the Frobenius norm, and $\mathcal{R}(\mathbf{U})$ is the regularized term with the parameter $\lambda'$. In addition, $\text{diag}(\cdot)$ denotes the diagonal elements in the matrix. We use the Orthogonal Matching Pursuit method (OMP) [Tropp and Gilbert, 2007] to learn the coefficients, since we have the prior knowledge of the number of samples in every cluster, which is the product of the number of sample sequences in each cluster and the size of the sequence. Then the affinity matrix is defined as $|\mathbf{U}| + |\mathbf{U}|^T$, where $|\cdot|$ is the absolute value. Ideally, this matrix is block diagonal given that samples in different $\mathbf{Y_i}$'s lie in independent affine subspaces. Finally, with the affinity matrix, the spectral clustering method is used to infer the clustering assignment $\hat{C}$. The clustering performance is measured by the normalized mutual information $\text{NMI}(\hat{C}, C)$ where $C$ is the ground truth division of $\mathbf{Y_i}$'s in $\mathbf{Y}$. The performance measures how well clusters of different variations can be separated, and thus it evaluates the disentanglement of factors. This is the first part of our measure.

The other part is the mean distance from the observed samples to the affine space spanned by generated samples. This measures the closeness described in the second assumption. Formally, given $x$ as an observed sample, we have the distance from $x$ to its projection on the affine space spanned by $\mathbf{Y}$ as

$$d(x, \mathbf{Y}) = \min_{\mathbf{u}} \|\mathbf{Yu} - \mathbf{x}\|_{\mathbf{2}}. \quad (13)$$

When all the data are normalized to have unit length, we have $d(x, \mathbf{Y}) \in [\mathbf{0}, \mathbf{1}]$. Given the observed set $X$ with $n$ samples, the averaged distance is $\bar{d} = \frac{1}{n} \sum_{x \in X} d(x, \mathbf{Y})$. We use $1 - \bar{d}$ to estimate the closeness.

Finally, our proposed subspace score is defined as

$$\alpha \, \text{NMI}(\hat{C}, C) + (1 - \alpha)(1 - \bar{d}). \quad (14)$$

We set $\alpha$ as 0.5 by default. The higher value of the subspace score is, the better the model learns the disentangled representation and generates new samples.

One may argue that the subspaces given by the principal component analysis (PCA) would get the highest score, because the assumptions of the subspace score are intrinsically compatible with PCA. However, the subspace score is to measure the variations of the generated samples. PCA lacks the ability to generate new samples, and it is not guaranteed to learn a concordant variation in each independent subspace. Thus, we do not think it is appropriate to apply the proposed metric on PCA. A novel disentanglement metric is also proposed in [Higgins *et al.*, 2017], which can measure the independence and interpretability of factors simultaneously. However, this metric is applied on a synthetic dataset of 2D shapes with predefined factors, but not on the real-world dataset.

## 4 Related Works

Several methods have been proposed for the unsupervised disentangled representation learning. InfoGAN [Chen *et al.*, 2016] designs an extra network to recover the code values from the generated samples to learn disentanglement. This is based on the principle of maximizing the mutual information, and thus InfoGAN shares the theoretical foundation with our strategy. The difference between our strategy and InfoGAN is that our methods recover the relative relation from samples. We believe that the relation between samples are more important than the code values, because the code values have little physical meaning. For example, we cannot infer how many pixels of the stroke width is of a generated digital number from the code. However, the comparison between the sample codes leads to the conclusion that one digit is bolder than the other. Thus, the analogical training of sample pairs is more conceptually straightforward.

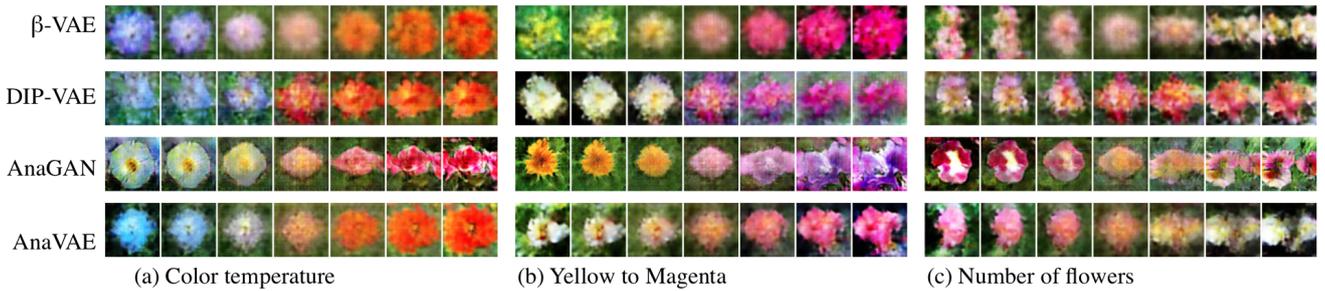

Figure 4: **Latent factors learned in the Flower dataset.** The pictures are generated by varying a specific latent variable from $-3$ to $3$ while others are zero. Each figure grid shows the variation of the similar factors, and each row shows the samples of the same method. The models learn to disentangle factors including the color temperature of the flower (a), the variation of the color from yellow to magenta (b), and the number of flowers (c). The pictures are best viewed magnified on screen.

Different from InfoGAN, $\beta$-VAE [Higgins *et al.*, 2017] is a variant of VAE. It learns the factorized factors by putting more emphasis on the KL-divergence between the variational posterior and the standard normal prior at the cost of the reconstruction accuracy. To avoid the cost, DIP-VAE [Kumar *et al.*, 2017] designs an extra regularization term which minimizes the KL-divergence between the expectation of the variation posterior over the data and the prior. The main difference between these methods and ours is the way to encourage disentanglement. The existing methods disentangle the factors by leveraging the statistical independence in the standard normal prior distribution. However, in a direct manner our analogical training strategy requires that the change of one factor does not result in the variations of the other factors in the generated samples. This constraint allows the model to learn the disentanglement of factors directly.

## 5 Experiments

In this section, we present the experiment results on five image datasets, including MNIST [LeCun *et al.*, 1998], CelebA [Liu *et al.*, 2015], Flower [Nilsback and Zisserman, 2008], CUB [Wah *et al.*, 2011] and Chairs [Aubry *et al.*, 2014]. Specifically, we compare our methods with other state-of-the-art methods along the subspace score.

### 5.1 Implementation Details

Implementation details in the experiments are summarized here. For AnaGAN, we use the network architecture of DCGAN [Radford *et al.*, 2015]. We use the WGAN-GP loss [Gulrajani *et al.*, 2017] instead of the original GAN loss and the parameter *num_critic* is set as 3. At the first 100 epoch, we only optimize $V(D, G)$, because at the beginning of the training $G$ has not captured the distribution yet and fails to generate samples good enough, and $R$ cannot learn useful relations from $G$. Both the noise and the code are sampled from the standard normal distribution. We use Adam optimizer [Kingma and Ba, 2014] with a learning rate of $0.00002$ and a momentum of $0.5$. The batch size is $32$. AnaVAE shares most of the configuration in AnaGAN. The encoder network $Q$ borrows the major structure of $D$ in AnaGAN. The learning rate for the Adam optimizer is $0.0001$. In the experiment on MNIST with labels, we combine all methods in the comparison with AC-GAN [Odena *et al.*, 2016] to incorporate the label information.

In both AnaVAE and AnaGAN, $R$ has the similar structure of $D$, but the numbers of feature mappings in convolutional layers are halved. When learning the analogical relation of the factor $r_i$, we first sample two identical codes $c_1$ and $c_2$. Then we set $c_{2i}$ (the $i$th component of $c_2$) as the $c_{1i} - v$ or $c_{1i} + v$ where $v$ is sampled from the uniform distribution over $[1, 2]$. If we take $c_{2i} = c_{1i} - v$ only, $R$ may separate a generative factor into two symmetric ones whose variations are in the opposite directions. The noise $z_1$ and $z_2$ are identical. We add a dropout layer [Srivastava *et al.*, 2014] after each nonlinear activation layer in $R$ to avoid overfitting. Lastly, the proposed algorithms are implemented with PyTorch.

To compute the subspace score, a cluster of ten sample sequences is generated for each factor and each sequence has five samples. The sequence is generated by varying the corresponding component of the code from $-2$ to $2$ with the interval $1$ but keeping other components fixed. We compute the subspace score over five different sets of generated samples to get the average. We implement it with scikit-learn.

### 5.2 Comparison

Figure 4 and Figure 5 provide a qualitative comparison of our proposed methods with $\beta$-VAE and DIP-VAE as appropriate. The pictures are generated by varying a specific latent variable from $-3$ to $3$ while others are fixed to zero, so the samples in the central column of each figure grid are the same. This guarantees that the effect of only one factor is investigated every time. Figure 4 shows the learned factors in the Flower dataset. The models learn the factors including the color temperature of the flower, the variation of the color from yellow to magenta, and the number of flowers automatically. The first two factors are concerned with the colors, while the last with the structure of the picture. The $\beta$-VAE and DIP-VAE tend to generate the blurry pictures, in which the petals and pistils are not distinguishable, while AnaVAE can generate flowers in a more clear fashion and AnaGAN can even generate flowers with detailed textures. Particularly, although all the methods learn the variations of the number of flowers in the data, $\beta$-VAE and DIP-VAE fail to represent the

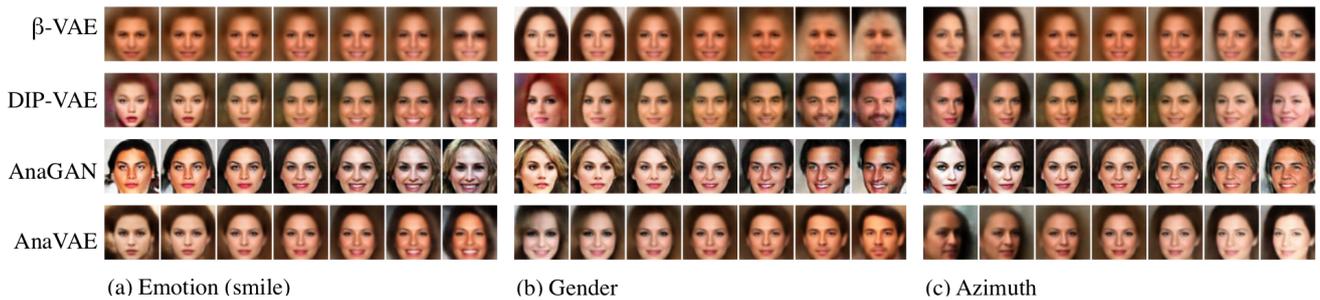

(a) Emotion (smile)      (b) Gender      (c) Azimuth

Figure 5: **Latent factors learned in the CelebA dataset.** The pictures are generated by varying a specific latent variable from $-3$ to 3 while others are zero. Each figure grid shows the variation of the similar factors, and each row shows the samples generated by the same method. The models learn to disentangle factors including the emotion (a), the transformation from female to male (b), and the azimuth (c). The pictures are best viewed magnified on screen.

Table 1: **Subspace score.** The best performances are highlighted.

| Dataset | MNIST (w/o labels) | CelebA | Flower | CUB | Chairs |
|---|---|---|---|---|---|
| VAE (untrained) [Kingma and Welling, 2013] | 0.208 / 0.417 | 0.095 | 0.132 | 0.145 | 0.085 |
| VAE [Kingma and Welling, 2013] | 0.552 / 0.608 | 0.546 | 0.551 | 0.555 | 0.565 |
| InfoGAN [Chen et al., 2016] | 0.389 / 0.482 | 0.484 | 0.445 | 0.452 | 0.198 |
| $\beta$-VAE ($\beta = 20$) [Higgins et al., 2017] | 0.508 / 0.582 | 0.512 | 0.510 | 0.508 | 0.538 |
| DIP-VAE ($\lambda = 10$) [Kumar et al., 2017] | **0.552 / 0.608** | 0.542 | 0.550 | 0.552 | 0.561 |
| AnaGAN (ours) | 0.403 / 0.495 | 0.541 | 0.478 | 0.491 | 0.536 |
| AnaVAE (ours) | 0.504 / 0.587 | **0.586** | **0.603** | **0.599** | **0.580** |

two flowers in one picture in a clean way, while our proposed method can clearly separate the two flowers. The color of the flowers here is entangled with the flower shapes, since we can seldom have a sunflower in magenta in reality.

Figure 5 shows the comparison in the CelebA dataset. The models are able to disentangle factors of emotion, gender, and azimuth of the faces. The $\beta$-VAE tends to give face in a blurry pattern, but it disentangles the factors in a relative explicit way. DIP-VAE and AnaVAE have similar clarity here, but the factors they learn are slightly entangled with other variations. Specifically, the factors of emotion they learn are related to the width of the face; while the faces become smiling from left to right, they also become fatter. At last, AnaGAN gives faces with much more details than the other methods, but it tends to entangle the factors with other variations. The first two factors are entangled with the hair color. Generally, our proposed methods show comparable performances with other methods.

Table 1 reports the subspace score of models. AnaVAE outperforms other methods in the experiments on most datasets, while DIP-VAE has the highest score on the MNIST dataset. VAE has a similar performance to DIP-VAE. It has been discussed in [Kumar et al., 2017] that VAE also has the ability to disentangle factors in the data. AnaGAN does not have high scores here. Since the subspace score uses the mean squared error to estimate reconstructions, it is hard for samples of AnaGAN with detailed textures to have high reconstruction accuracies. InfoGAN also suffers from the same difficulty. In summary, AnaVAE shows competitive performances as compared with other methods along the subspace score.

## 6 Conclusion

In this paper, we propose the analogical training strategy to learn the disentangled representation without supervision. Based on the observation that there exists an analogical relation unique to a generative factor, the proposed strategy learns the disentangled factors from the generated analogical sample pairs. The strategy designs a classifier to help the generator to generate sample pairs of consistent analogical relations. We also propose the subspace score to measure the disentanglement of the factors. The subspace score does not require label information in the dataset. Experiments show that our proposed methods can uncover interpretable factors in the data and give competitive performances as compared with other state-of-the-art algorithms. In the future work, we will explore different measures of disentanglement and the application of the disentangled representation in foundational AI problems, such as the conceptual space learning.

## Acknowledgments

This work is funded by the National Natural Science Foundation of China (NSFC) under Grant NO. 61773336. We would like to extend our gratitude to the data providers. Zejian Li would like to thank Wei Li, Chao Wang and Ting He in Zhejiang University for helpful comments.

## References

[Aubry et al., 2014] Mathieu Aubry, Daniel Maturana, Alexei A. Efros, Bryan C. Russell, and Josef Sivic. Seeing 3d chairs: exemplar part-based 2d-3d alignment using a large dataset of cad models. In *Proceedings of the IEEE conference on computer vision and pattern recognition*, pages 3762–3769. IEEE, 2014.

[Bechberger and Kühnberger, 2017] Lucas Bechberger and Kai-Uwe Kühnberger. Towards grounding conceptual spaces in neural representations. *arXiv preprint arXiv:1706.04825*, 2017.

[Bengio et al., 2013] Yoshua Bengio, Aaron Courville, and Pascal Vincent. Representation learning: A review and new perspectives. *IEEE transactions on pattern analysis and machine intelligence*, 35(8):1798–1828, 2013.

[Chen et al., 2016] Xi Chen, Yan Duan, Rein Houthooft, John Schulman, Ilya Sutskever, and Pieter Abbeel. Infogan: Interpretable representation learning by information maximizing generative adversarial nets. In *Advances in Neural Information Processing Systems*, pages 2172–2180, 2016.

[Denton and Birodkar, 2017] Emily Denton and Vighnesh Birodkar. Unsupervised learning of disentangled representations from video. In *Advances in Neural Information Processing Systems*, pages 4417–4426. Curran Associates, Inc., 2017.

[Derrac and Schockaert, 2015] Joaquín Derrac and Steven Schockaert. Inducing semantic relations from conceptual spaces: a data-driven approach to plausible reasoning. *Artificial Intelligence*, 228:66–94, 2015.

[Dosovitskiy et al., 2015] Alexey Dosovitskiy, Jost Tobias Springenberg, and Thomas Brox. Learning to generate chairs with convolutional neural networks. In *Proceedings of the IEEE Conference on Computer Vision and Pattern Recognition*, pages 1538–1546. IEEE, 2015.

[Elhamifar and Vidal, 2013] Ehsan Elhamifar and Rene Vidal. Sparse subspace clustering: Algorithm, theory, and applications. *IEEE transactions on pattern analysis and machine intelligence*, 35(11):2765–2781, 2013.

[Gärdenfors, 2004] Peter Gärdenfors. *Conceptual spaces: The geometry of thought*. MIT press, 2004.

[Goodfellow et al., 2014] Ian Goodfellow, Jean Pouget-Abadie, Mehdi Mirza, Bing Xu, David Warde-Farley, Sherjil Ozair, Aaron Courville, and Yoshua Bengio. Generative adversarial nets. In *Advances in neural information processing systems*, pages 2672–2680, 2014.

[Gulrajani et al., 2017] Ishaan Gulrajani, Faruk Ahmed, Martin Arjovsky, Vincent Dumoulin, and Aaron Courville. Improved training of wasserstein gans. *arXiv preprint arXiv:1704.00028*, 2017.

[Gust et al., 2008] Helmar Gust, Ulf Krumnack, Kai-Uwe Kühnberger, and Angela Schwering. Analogical reasoning: A core of cognition. In *Proceedings of the 31st Annual German Conference on AI, KI 2008*, volume 22, pages 8–12. Springer, 2008.

[Higgins et al., 2017] Irina Higgins, Loic Matthey, Arka Pal, Christopher Burgess, Xavier Glorot, Matthew Botvinick, Shakir Mohamed, and Alexander Lerchner. B-vae: Learning basic visual concepts with a constrained variational framework. In *International Conference on Learning Representation*, 2017.

[Ioffe and Szegedy, 2015] Sergey Ioffe and Christian Szegedy. Batch normalization: Accelerating deep network training by reducing internal covariate shift. In *Proceedings of the 32nd International Conference on Machine Learning*, volume 37, pages 448–456. PMLR, 2015.

[Kingma and Ba, 2014] Diederik Kingma and Jimmy Ba. Adam: A method for stochastic optimization. *arXiv preprint arXiv:1412.6980*, 2014.

[Kingma and Welling, 2013] Diederik P. Kingma and Max Welling. Auto-encoding variational bayes. *arXiv preprint arXiv:1312.6114*, 2013.

[Kulkarni et al., 2015] Tejas D. Kulkarni, William F. Whitney, Pushmeet Kohli, and Joshua B. Tenenbaum. Deep convolutional inverse graphics network. In *Advances in Neural Information Processing Systems*, pages 2539–2547. Curran Associates, Inc., 2015.

[Kumar et al., 2017] Abhishek Kumar, Prasanna Sattigeri, and Avinash Balakrishnan. Variational inference of disentangled latent concepts from unlabeled observations. *arXiv preprint arXiv:1711.00848*, 2017.

[Lawry and Tang, 2009] Jonathan Lawry and Yongchuan Tang. Uncertainty modelling for vague concepts: A prototype theory approach. *Artificial Intelligence*, 173(18):1539–1558, 2009.

[LeCun et al., 1998] Yann LeCun, Léon Bottou, Yoshua Bengio, and Patrick Haffner. Gradient-based learning applied to document recognition. *Proceedings of the IEEE*, 86(11):2278–2324, 1998.

[Lewis and Lawry, 2016] Martha Lewis and Jonathan Lawry. Hierarchical conceptual spaces for concept combination. *Artificial Intelligence*, 237:204–227, 2016.

[Liu et al., 2015] Ziwei Liu, Ping Luo, Xiaogang Wang, and Xiaoou Tang. Deep learning face attributes in the wild. In *Proceedings of International Conference on Computer Vision (ICCV)*, pages 3730–3738. IEEE, 2015.

[Nilsback and Zisserman, 2008] Maria-Elena Nilsback and Andrew Zisserman. Automated flower classification over a large number of classes. In *Indian Conference on Computer Vision, Graphics & Image Processing, 2008.*, pages 722–729. IEEE, 2008.

[Odena et al., 2016] Augustus Odena, Christopher Olah, and Jonathon Shlens. Conditional image synthesis with auxiliary classifier gans. *arXiv preprint arXiv:1610.09585*, 2016.

[Peng et al., 2017] Xi Peng, Xiang Yu, Kihyuk Sohn, Dimitris N Metaxas, and Manmohan Chandraker. Reconstruction-based disentanglement for pose-invariant face recognition. In *Proceedings of the IEEE International Conference on Computer Vision*, pages 1632–1641, 2017.

[Radford *et al.*, 2015] Alec Radford, Luke Metz, and Soumith Chintala. Unsupervised representation learning with deep convolutional generative adversarial networks. *arXiv preprint arXiv:1511.06434*, 2015.

[Srivastava *et al.*, 2014] Nitish Srivastava, Geoffrey Hinton, Alex Krizhevsky, Ilya Sutskever, and Ruslan Salakhutdinov. Dropout: a simple way to prevent neural networks from overfitting. *Journal of machine learning research*, 15(1):1929–1958, 2014.

[Tang and Xiao, 2017] Yongchuan Tang and Yunsong Xiao. Learning fuzzy semantic cell by principles of maximum coverage, maximum specificity, and maximum fuzzy entropy of vague concept. *Knowledge-Based Systems*, 2017.

[Tropp and Gilbert, 2007] Joel A. Tropp and Anna C. Gilbert. Signal recovery from random measurements via orthogonal matching pursuit. *IEEE Transactions on Information Theory*, 53(12):4655–4666, 2007.

[Ulyanov *et al.*, 2016] Dmitry Ulyanov, Andrea Vedaldi, and Victor Lempitsky. Instance normalization: The missing ingredient for fast stylization. *arXiv preprint arXiv:1607.08022*, 2016.

[Wah *et al.*, 2011] Catherine Wah, Steve Branson, Peter Welinder, Pietro Perona, and Serge Belongie. The caltech-ucsd birds-200-2011 dataset. 2011.

[Wang *et al.*, 2017] Chaoyue Wang, Chaohui Wang, Chang Xu, and Dacheng Tao. Tag disentangled generative adversarial network for object image re-rendering. In *Proceedings of the Twenty-Sixth International Joint Conference on Artificial Intelligence, IJCAI*, pages 2901–2907, 2017.

[Worrall *et al.*, 2017] Daniel E. Worrall, Stephan J. Garbin, Daniyar Turmukhambetov, and Gabriel J. Brostow. Interpretable transformations with encoder-decoder networks. In *The IEEE International Conference on Computer Vision (ICCV)*, volume 4, 2017.

## Appendix A

This part is to give more details of the experiment setup.

We begin with the preprocessing of the data. The images of CelebA, Flower, CUB, and Chairs are cropped and rescaled to the size of $64 \times 64$. The images of all datasets are normalized to the range of $[0, 1]$ for AnaVAE and $[-1, 1]$ for AnaGAN. For the design of the model, we use the transposed convolution operation [Dosovitskiy *et al.*, 2015] in the generative network as suggested in DCGAN [Radford *et al.*, 2015]. We apply batch normalization [Ioffe and Szegedy, 2015] after most layers. For the discriminator in AnaGAN, we apply instance normalization [Ulyanov *et al.*, 2016] instead of layer normalization suggested in [Gulrajani *et al.*, 2017]. For the analogical relation classifier $R$, we add a Dropout layer [Srivastava *et al.*, 2014] with a dropout rate of $0.2$ after each nonlinear activation layer. We use the softmax layer as the last layer to perform the classification. Detailed architectures are shown in tables 2 to 5. We use some abbreviations here. We denote "Conv." as the convolution layer and UpConv as the transposed convolution layer. For a convolution layer, "4c2s64o" means we use the $4 \times 4$ kernel with the stride as 2, and the output number of feature mapping is 64. The transposed convolution layer shares the same notation. We denote rectified linear unit as ReLU and the leaky linear unit as LReLU. "BN" means the batch normalization layer and "FC" means the fully connected layer.

In the training of AnaGAN, we apply the trick that at the first 25 iteration of the generator $G$ the *num_critic* is set as 100 so as to initialize the discriminator well enough. The regularization parameter of the gradient penalty is $0.25$. We use the ADAM [Kingma and Ba, 2014] optimizer with $\beta_1 = 0.5$ and $\beta_2 = 0.99$.

To compute the subspace score, a cluster of ten sample sequences is generated for each factor and each sequence has five samples. The sequence is generated by varying the corresponding component of the code from $-2$ to $2$ with the interval $1$ but keeping other components fixed. The coefficients in the subspace clustering is learned via the Orthogonal Matching Pursuit [Tropp and Gilbert, 2007]. In the computation of the distance from the observed to the generated affine subspace, we only sample $6,400$ observed samples. The subspace score is computed over five different sets of generated samples to get the average.

## Appendix B

In this part, we discuss the potential applications of our proposed analogical learning strategy.

The analogical learning strategy can be applied in supervisedly or semi-supervisedly disentangled factors learning, too. Given that we have sample pairs $x_1$ and $x_2$ and we know from the ground truth that they are different in the generative factor $r$, we can train $R$ with this supervised signal via the following optimization.

$$\max_{G,R} K_{sup}(G, R) = \sum_{(x_1, x_2), r} \log R(r \mid x_1, x_2) \quad (15)$$

This is to maximize the empirical mutual information between $x_1$, $x_2$ and $r$ directly without the generating process. Such additional supervised training is compatible with the current framework. In semi-supervised learning case, our proposed methods can incorporate this supervised signal to learn the given generative factors and other unknown independent factors. However, such pairwise constraints are expensive to get in the real-world data. Thus, this extension may only be applied in the learning of some predefined transformations of data, such as rotation or scaling.

AnaVAE has the potential application in learning the conceptual space in uncertainty modeling. Traditionally, the conceptual space consists of quality dimensions which specify some attributes of the concept. A typical example is the RGB color space in which each color is defined by the linear combination of three color elements. Since the disentangled factors learned by AnaVAE can be interpretable, the latent code space is an approximation of the conceptual space. Simultaneously, the input noise can be used to model the uncertainty of each concept, and the generator provides a mapping from the conceptual space to the real-world feature space to visualize the concepts. In this scenario, the prior distribution of the codes and the noises can be defined as some other well-studied structure in the literature of cognitive science [Gärdenfors, 2004], such as the convex structures based on the prototype theory [Lawry and Tang, 2009] or other variants [Lewis and Lawry, 2016; Tang and Xiao, 2017]. A recent exciting preprint [Bechberger and Kühnberger, 2017] discusses the possibility to learn conceptual space with InfoGAN [Chen *et al.*, 2016]. Another approach [Derrac and Schockaert, 2015] also learns the space via embedding techniques and defines the salient direction (the quality dimension) by fitting linear Support Vector Machine model on the data. In the learned space they can find analogical relations and other semantic relations. AnaVAE catches some principles of the methods above and may have the potential to learn a conceptual space, which captures the quality dimension by the analogical relations. We will explore this in our future work.

Table 2: **AnaVAE for MNIST.** We use a code size of 16.

| Encoder Q | Generator G | Classifier R |
|---|---|---|
| Input $1 \times 28 \times 28$ gray images | Input code $\in \mathbb{R}^{16}$ | Input $2 \times 28 \times 28$ gray images |
| 4c2s64o Conv. LReLU(0.2) | 1024 FC. BN. ReLU. | 4c2s32o Conv. LReLU(0.2) Dropout(0.2) |
| 4c2s128o Conv. BN. LReLU(0.2) | $128 \times 7 \times 7$ FC. BN. ReLU. | 4c2s64o Conv. BN. LReLU(0.2) Dropout(0.2) |
| 1024 FC. BN. LReLU(0.2) | 4c2s64o UpConv. BN. ReLU. | 1024 FC. BN. LReLU(0.2) Dropout(0.2) |
| 32 FC. | 4c2s1o UpConv. Sigmoid. | 16 FC. Softmax. |

Table 3: **AnaGAN for MNIST.** We use a code size of 8 and a noise size of 16.

| Discriminator D | Generator G | Classifier R |
|---|---|---|
| Input $1 \times 28 \times 28$ gray images | Input code $\in \mathbb{R}^{24}$ | Input $2 \times 28 \times 28$ gray images |
| 4c2s64o Conv. LReLU(0.2) | 1024 FC. BN. ReLU. | 4c2s32o Conv. LReLU(0.2) Dropout(0.2) |
| 4c2s128o Conv. IN. LReLU(0.2) | $128 \times 7 \times 7$ FC. BN. ReLU. | 4c2s64o Conv. BN. LReLU(0.2) Dropout(0.2) |
| 1024 FC. IN. LReLU(0.2) | 4c2s64o UpConv. BN. ReLU. | 1024 FC. BN. LReLU(0.2) Dropout(0.2) |
| 1 FC. | 4c2s1o UpConv. Tanh. | 8 FC. Softmax. |

Table 4: **AnaVAE for CelebA, Flower, CUB, and Chairs.** We use a code size of 32 for the CelebA and Chairs dataset and 64 for the Flower and CUB dataset.

| Encoder Q | Generator G | Classifier R |
|---|---|---|
| Input $3 \times 64 \times 64$ color images | Input code $\in \mathbb{R}^{32}$ (CelebA and Chairs) | Input $6 \times 64 \times 64$ color images |
| | Input code $\in \mathbb{R}^{64}$ (Flower and CUB) | |
| 4c2s64o Conv. LReLU(0.1) | 4c1s512o UpConv. BN. LReLU(0.1). | 4c2s32o Conv. LReLU(0.1) Dropout(0.2) |
| 4c2s128o Conv. BN. LReLU(0.1) | 4c2s256o UpConv. BN. LReLU(0.1). | 4c2s64o Conv. BN. LReLU(0.1) Dropout(0.2) |
| 4c2s256o Conv. BN. LReLU(0.1) | 4c2s128o UpConv. BN. LReLU(0.1). | 4c2s128o Conv. BN. LReLU(0.1) Dropout(0.2) |
| 4c2s512o Conv. BN. LReLU(0.1) | 4c2s64o UpConv. BN. LReLU(0.1). | 4c2s256o Conv. BN. LReLU(0.1) Dropout(0.2) |
| 4c1s64o Conv. (CelebA and Chairs) | 4c2s3o UpConv. Sigmoid. | 4c1s32o Conv. Softmax. (CelebA and Chairs) |
| 4c1s128o Conv. (Flower and CUB) | | 4c1s64o Conv. Softmax. (Flower and CUB) |

Table 5: **AnaGAN for CelebA, Flower, CUB, and Chairs.** We use a code size of 32 and a noise size of 64.

| Discriminator D | Generator G | Classifier R |
|---|---|---|
| Input $3 \times 64 \times 64$ color images | Input code $\in \mathbb{R}^{96}$ | Input $6 \times 64 \times 64$ color images |
| 4c2s64o Conv. LReLU(0.1) | 4c1s512o UpConv. BN. LReLU(0.1). | 4c2s32o Conv. LReLU(0.1) Dropout(0.2) |
| 4c2s128o Conv. IN. LReLU(0.1) | 4c2s256o UpConv. BN. LReLU(0.1). | 4c2s64o Conv. BN. LReLU(0.1) Dropout(0.2) |
| 4c2s256o Conv. IN. LReLU(0.1) | 4c2s128o UpConv. BN. LReLU(0.1). | 4c2s128o Conv. BN. LReLU(0.1) Dropout(0.2) |
| 4c2s512o Conv. IN. LReLU(0.1) | 4c2s64o UpConv. BN. LReLU(0.1). | 4c2s256o Conv. BN. LReLU(0.1) Dropout(0.2) |
| 4c1s1o Conv. | 4c2s3o UpConv. Tanh. | 4c1s32o Conv. Softmax. |